\documentclass[final]{cvpr}

\usepackage{times}
\usepackage{epsfig}
\usepackage{graphicx}
\usepackage{amsmath}
\usepackage{amssymb}

\usepackage{caption}
\usepackage{subcaption}
\usepackage{multicol}
\usepackage{multirow}

\usepackage[pagebackref=true,breaklinks=true,colorlinks,bookmarks=false]{hyperref}

\def\cvprPaperID{4079} % *** Enter the CVPR Paper ID here
\def\confYear{CVPR 2021}

\begin{document}

%%%%%%%%% TITLE
\title{AET-EFN: A Versatile Design for Static and Dynamic Event-Based Vision}

\author{Chang Liu, Xiaojuan Qi, Edmund Lam, Ngai Wong\\
The University of Hong Kong\\
% Pokfulam Road, Hong Kong\\
{\tt\small lcon7@connect.hku.hk \{xjqi, elam, nwong\}@eee.hku.hk}
\and
}

\maketitle

%%%%%%%%% ABSTRACT
\begin{abstract}
   The neuromorphic event cameras, which capture the optical changes of a scene, have drawn increasing attention due to their high speed and low power consumption. 
However, the event data are noisy, sparse, and nonuniform in the spatial-temporal domain with extremely high temporal resolution, making it challenging to design backend algorithms for event-based vision. Existing methods encode events into point-cloud-based or voxel-based representations, but suffer from noise and/or information loss. Additionally, there is little research that systematically studies how to handle static and dynamic scenes with one universal design for event-based vision. 
This work proposes the Aligned Event Tensor (AET) as a novel event data representation, and a neat framework called Event Frame Net (EFN), which enables our model for event-based vision under static and dynamic scenes. The proposed AET and EFN are evaluated on various datasets, and proved to surpass existing state-of-the-art methods by large margins. Our method is also efficient and achieves the fastest inference speed among others.
\end{abstract}

%%%%%%%%% BODY TEXT
\section{Introduction}

% what is event camera, and its advantage
Recently, the neuromorphic event cameras have drawn increasing attention from researchers. Unlike traditional cameras, which take a whole frame periodically, event cameras asynchronously capture the per-pixel optical changes instead of the absolute brightness. Event cameras enjoy the advantages of low latency (in microsecond level), high dynamic range, low power consumption, and low bandwidth cost \cite{lichtsteiner2008128}. Because of the appealing characteristics of event cameras, event-based vision 
is gaining more attention in many research areas and applications, such as robotics, image deblurring and auto-driving \cite{survery}.

% concise intro of event cam, and the need for new algorithms
The outputs of event cameras are event flows defined by $(x, y, t, p)$, which are pixel coordinates, timestamp, and polarity. The event flows denote instantaneous increases or decreases of the light intensity, and are incompatible with typical practices such as convolutional neural networks (CNNs). Besides the unique data structure, event data are challenging to process also because they do not contain information about the absolute intensity but many noises. Therefore, to fully leverage the advantages, new algorithms that can efficiently process event data are highly desired~\cite{survery}.

\begin{figure}
    \centering
    \begin{subfigure}[]{0.95\columnwidth}
        \centering
        \includegraphics[width=\linewidth]{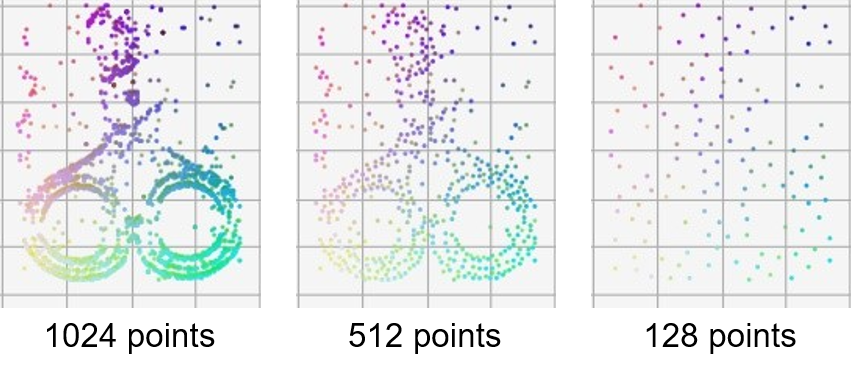}
        \vspace{-20pt}
        \caption{Downsampling Event Data with FPS algorithm. The outlier noises are kept during downsampling and the shape collapses.}
        \label{fig:noises}
    \end{subfigure}
    \vspace{5pt}
    
    \begin{subfigure}[]{0.95\columnwidth}
        \centering
        \begin{subfigure}[]{0.32\linewidth}
        \includegraphics[width=\linewidth]{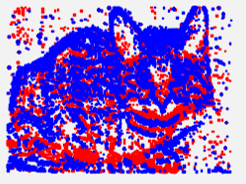}
        \end{subfigure}
        \begin{subfigure}[]{0.32\columnwidth}
            \includegraphics[width=\linewidth]{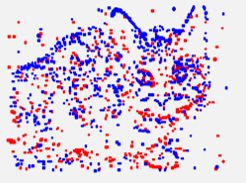}
        \end{subfigure}
        \begin{subfigure}[]{0.32\columnwidth}
            \includegraphics[width=\linewidth]{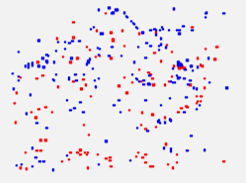}
        \end{subfigure}
        \vspace{-5pt}
        \caption{Events accumulated for 1/50, 1/100, 1/500 total time. Proper duration of accumulation is needed for clear image.}
        \label{fig:quantize_vis}
    \end{subfigure}
    
    \vspace{-5pt}
    \caption{Visualization of event data samples.}
    \vspace{-15pt}
    \label{fig:intro_fig}
\end{figure}

% existing works, their pros and cons.
To leverage event cameras for vision tasks, some existing approaches regard the event data as point clouds~\cite{yang2019gumbel,qi2017pointnet}, and directly apply point-cloud-based models. However, the point-cloud-based methods such as \cite{qi2017pointnet, liu2019relation} originally handle small clouds with around 1024 points, hence are not efficient for large clouds such as event data with hundreds of thousands of events per second. Moreover, the downsampling algorithms, such as the Furthest Point Sampling used in the point-cloud-based methods~\cite{qi2017pointnet++,liu2019relation}, are prone to noises commonly found in event data, as shown in Fig.~\ref{fig:noises}. Consequently, point-cloud-based methods have limitations in accuracy and efficiency for event-based vision.

Besides point-cloud-based methods, the voxel-based approaches compress the event data into voxels, bridging event cameras with conventional CNN techniques \cite{est, lagorce2016hots, sironi2018hats, cannici2020matrix}. The voxel-based approaches such as \cite{est} and \cite{cannici2020matrix} achieve the state-of-the-art (SOTA) performance on several event-based datasets, while enjoying higher efficiency than point-cloud-based methods, due to the good regularity for parallel computing on GPUs.

\label{sec:intro}
Nonetheless, the voxel-based methods have to compress the data drastically to meet the memory and computational requirements, and hence suffer from information loss. According to the observation in our early-stage experiments, the reduced temporal resolution also causes problems such as motion blur and accuracy drop for dynamic scenes. Moreover, to fit the neural networks originally designed for image data, they usually stack the different frames along the channel axis~\cite{est, cannici2020matrix}, which makes the temporal information tangle with channel information. Therefore, voxel-based methods are limited by information loss and flexibility under scenes at different speeds.

% Our approach, motivation and the gaps we fill
To this end, we propose an efficient representation for event data, which can alleviate the information loss problem of voxel-based methods, while also being robust under both static and dynamic scenes. We name it Aligned Event Tensor (AET). In AET, the event data are quantized and voxelized in high temporal resolution to minimize information loss. The voxels are then compressed in local spatial-temporal regions by convolutional layers to alleviate the misalignment problem across different frames. We also present a two-branch neural network design, the Event Frame Net (EFN), which is versatile for static and dynamic event-based vision. In EFN, a CNN encodes the AET frames into vector representations, which are then processed independently for frame-level predictions, or as a whole sequence for video-level prediction. 

Both AET and EFN can significantly increase the prediction accuracy for event-based vision in the experiments. They are lightweight and enjoy the fastest inference speed among the fellows. Our main contributions are:
\begin{itemize}
    \item We propose the efficient AET representation for event data, which improves speed and accuracy for event data classification tasks under different speed scenes.
    \item The Event Frame Net (EFN) can process event data under static and dynamic scenes, achieving the SOTA performances in all classification experiments.
    \item Detailed experiments and speed analysis are conducted. Our method significantly improves the accuracy while being the fastest among existing methods. Visualization also empirically justifies our design.
\end{itemize}

\section{Related Works}
% Reconstructive approaches, heavy, slow, costly (pretrain), simulated data
\subsection{Reconstructive approaches}
Several studies such as E2VID and EventSR~\cite{e2vid, wang2020eventsr} reconstruct events back to normal images for conventional vision. The reconstruction methods usually utilize recursive modules such as ConvLSTM~\cite{xingjian2015convLSTM} or structures like UNet~\cite{ronneberger2015UNet} to reconstruct verisimilar images based on the events \cite{e2vid, scheerlinck2020fast}. However, the massive pretraining workload is costly, and the cumbersome two-stage framework is limited for real-time applications. It is also challenging to get simultaneous ground-truth frames in the real world for training the reconstructors. Simulators like the ESIM~\cite{rebecq2018esim} are used to generate events from normal videos \cite{e2vid}, which still differ from real-world data. Due to these challenges, end-to-end models are also widely studied for downstream tasks (e.g., classification) \cite{est, cannici2020matrix}, where decisions are more interested than the frames.

% Point-cloud-based approaches, limited cloud size, prone to noises
\subsection{Point-cloud-based methods}
Among the end-to-end methods, some researchers tend to consider event data as 3D point clouds in the spatial-temporal space, and directly apply point-cloud networks. These point-cloud-based methods, such as \cite{qi2017pointnet, qi2017pointnet++, zhao2019pointweb, liu2019relation}, are originally designed to process spatial 3D point clouds with thousands of points. The works in~\cite{qi2017pointnet++},~\cite{zhao2019pointweb} and~\cite{liu2019relation} gradually aggregate local neighboring points into centroids, hence extracting comprehensive information from the point cloud. The schemes in~\cite{yang2019gumbel} and~\cite{wang2019space} adopt the point cloud networks for event data classification and demonstrate good results on the gesture classification dataset DVS128~\cite{dvs128}. 

However, point-cloud-based methods are usually limited by the scale of input clouds and are prone to noise. The event data contain hundreds of thousands of events per second, which are much larger than typical point clouds studied by the above methods (e.g., 1024 points). Therefore, the neighbor querying algorithms in the point-cloud-based models, such as K-Nearest-Neighbours \cite{yang2019gumbel} or Ball Query \cite{liu2019relation}, cannot process the event clouds efficiently due to the $O(N^2)$ complexity. Moreover, point cloud downsampling algorithms such as Furthest-Point-Sampling \cite{qi2017pointnet++} are prone to outlier noise in event data, as illustrated in Fig.~\ref{fig:noises}. For these reasons, point-cloud-based methods are not widely used for event-based vision.

% Voxel-based methods, memory constraints, fast, information loss, blur (misalignment)
\subsection{Voxel-based methods}
Voxel-based methods transform the event data into voxel form representations, which are compatible with conventional CNN techniques. The representations are either end-to-end trainable or based on heuristics. Two pioneering works are in~\cite{sironi2018hats} and~\cite{lagorce2016hots} that present the time-surface concept to convert the sparse events into a 2D representation. Later the Event Spike Tensor (EST)~\cite{est} assigns each event some weights, generated by a multilayer perceptron (MLP) kernel, and projects the events to different anchor frames. The EST is a milestone work for event-based vision and achieves SOTA performance at its time.

Inspired by~\cite{est}, Matrix-LSTM~\cite{cannici2020matrix} proposes an encoding process with paralleled long short-term memory (LSTM\cite{lstm}) cells. The history of events that have happened over each pixel is encoded into a vector by the LSTM, hence generating a multi-channel image. It improves the prediction accuracy but has a slower speed than EST. 

However, for dynamic scenarios, such as human activity recognition, the above approaches encounter motion blur problems due to the drastic compression in the compromise of memory cost (e.g., 30k times compression to 9 frames). The accuracy also drops because of information loss during the compression from spatial-temporal 3D to 2D. In addition, we argue that since events are triggered by the moving objects in the real world, the pixel-wise operation in~\cite{est, cannici2020matrix} will mistakenly compress events from different objects (or parts) into one element, causing the misalignment problem.

\section{Methodology}
% Introductory paragaph, signaling the following contents
In this section, we will introduce the proposed Aligned Event Tensor (\textbf{AET}) and the Event Frame Net (\textbf{EFN}). AET is an efficient voxel-based representation for event data, which is end-to-end trainable and alleviates problems of information loss, motion blur and misalignment, while EFN is a versatile network for handling both static and dynamic scenes for event-based vision.

\begin{figure*}[ht]
    \begin{center}
        \includegraphics[scale=0.4]{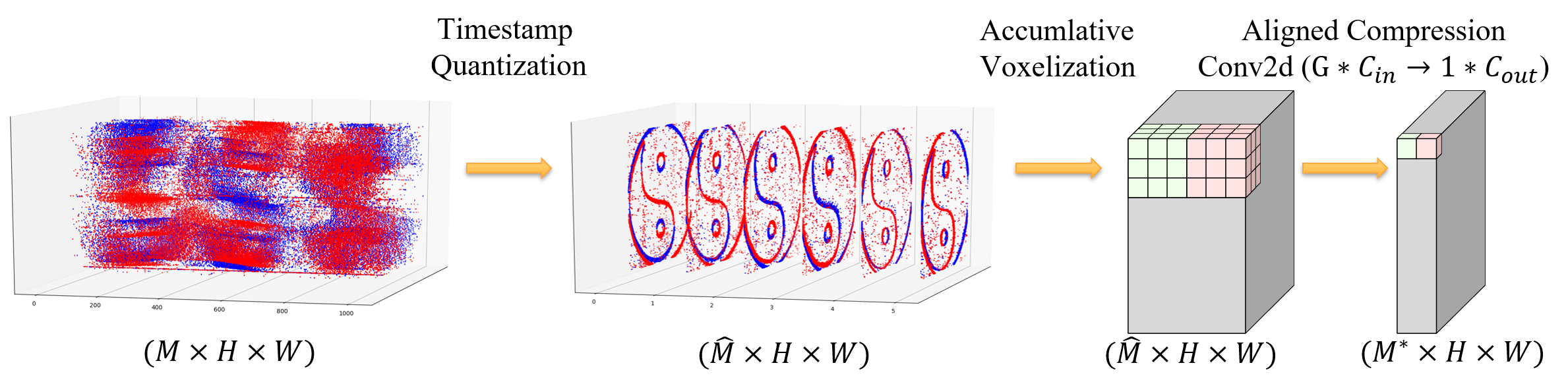}
        \vspace{-5pt}
        \caption{Aligned Event Tensor workflow. For a better visual effect, $\hat{M}=6$ and one Conv layer with $G=3$ are shown in this example. The events are quantized along the time axis, and then accumulatively voxelized (each voxel's value is the sum of all historical events). After that, the frames are grouped and compressed with convolutional layers.}
        \vspace{-15pt}
        \label{fig:AET_flow}
    \end{center}
\end{figure*}

\subsection{Aligned Event Tensor}
AET comprises three parts, namely, Timestamp Quantization, Accumulative Voxelization and Aligned Compression, as shown in Fig.~\ref{fig:AET_flow}.

\subsubsection{Timestamp Quantization}
\label{sec:TimeQuant}
% Motivation: Time resolution too high, must quantize for memory limitation
Event data are spatial-temporal points with an extremely high temporal resolution, which theoretically implies a full resolution tensor of shape $M\times H \times W$, where $M$ is the number of virtual frames, usually in the order of millions. Due to the large size and high sparsity, it is inefficient to process the whole tensor directly.

% Justify, observation: quantize with proper duration is fine
We also observed that accumulation of events during a proper period is needed to form a recognizable shape. The duration can neither be too short nor too long, otherwise the shape would be incomplete or blurred, as in Fig.~\ref{fig:quantize_vis}.

% Method: quantize time stamp into #bins
Therefore, we propose to quantize the timestamp with a properly high temporal resolution $\hat{M}$ (e.g., 100) to meet the computational resource constraints. Although the optimal value for $\hat{M}$ could be sample-varying, to facilitate batch training, a uniform time resolution is adopted for simplicity. For each sample, the timestamps are normalized into the range of $[0, 1]$, and then evenly quantized into $\hat{M}$ bins:

\vspace{-10pt}

\begin{equation}
\label{eqn:quantizeT}
    Q(\mathcal{T}_i^{(j)}) = \max\left(\left\lceil \frac{\hat{M} \times \left(\mathcal{T}_i^{(j)} - \min(\mathcal{T}_i)\right)}{\max(\mathcal{T}_i)-\min(\mathcal{T}_i)}  \right\rceil, 1 \right)
\end{equation}
where $\mathcal{T}_i$ is the set of timestamps of the $i$th sample's events and the superscript $j$ corresponds to the $j$th event.

By Timestamp Quantization, events with adjacent timestamps are aggregated together to form a recognizable shape. The high temporal resolution also alleviates information loss and motion blur than low-resolution methods such as EST~\cite{est} (9 frames). The procedure is in the left of Fig.~\ref{fig:AET_flow}.

\begin{figure}
    \centering
    \includegraphics[width=0.9\columnwidth]{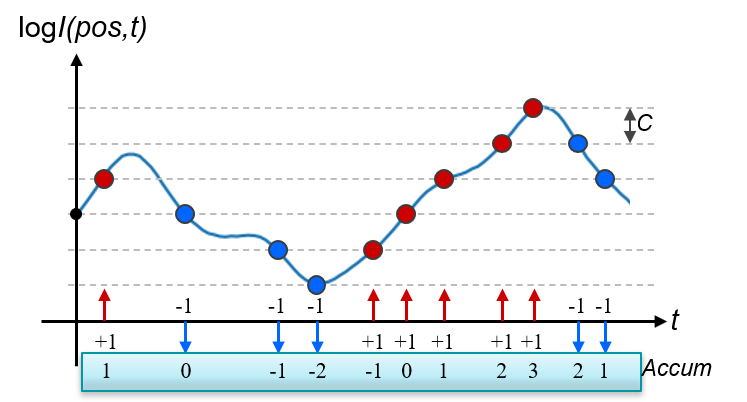}
    \vspace{-5pt}
    \caption{Optical intensity curve, with the triggered events. The accumulated values are shown in the bottom blue box.}
    \vspace{-15pt}
    \label{fig:eventCam_m}
\end{figure}

\subsubsection{Accumulative Voxelization}
\label{sec:accumulative_voxelization}
% Details about our method
After quantization, the events now imply an $\hat{M}\times H\times W$ tensor, mapped by their coordinates and quantized timestamps. Inspired by the intrinsic mechanism of the event cameras shown in Fig.~\ref{fig:eventCam_m}, we design the Accumulative Voxelization method to determine the value of each voxel. Recall that event cameras track optical intensity per pixel and generate a positive or negative event whenever the change exceeds the threshold, the summation of events at each position implies an absolute intensity change. At voxel $(m, y, x)$, we sum the past events by their polarities:
\begin{equation}
    \mathcal{V}(m,y,x) = \sum_{j}{\mathbb{I}(x_j=x, y_j=y, \tau_j \leq m)p_j}
\end{equation}
where $\mathcal{V}$ is the intermediate voxel representation of the events, $\tau_j = Q(\mathcal{T}_i^{(j)})$ is the quantized timestamp defined in Equation (\ref{eqn:quantizeT}), $\mathbb{I}$ is the indicator function and $p_j \in \{-1,+1\}$ is the polarity of the $j$th event. After the Accumulative Voxelization, $\mathcal{V} \in \mathbb{R}^{(1\times)\hat{M}\times H \times W}$ now denotes $\hat{M}$ frames with $C=1$ channels each.

% Justification: track long history, record absolute contrast (intrinsic event cam mechanism)
The Accumulative Voxelization method efficiently tracks a long history of the absolute intensity changes. Previous methods such as EST \cite{est} and Matrix-LSTM \cite{cannici2020matrix} use trainable kernels to encode histories. But they are more computationally intensive and memory consuming, thus not feasible for high temporal resolution as in AET, resulting in information loss. Moreover, in our method, we record the rough intensity changes with regard to the first frame, by summing positive and negative events together. Compared with EST alike methods, the moving traces fades more directly. The frames in $\mathcal{V}$ will be compressed in next step.

\subsubsection{Aligned Compression via Conv2D}
\label{sec:aligned_compress}
% Motivation: Compress frames for CNN to process, also address the misalignment problem
During quantization and voxelization, the temporal resolution $\hat{M}$ is kept high to alleviate information loss. However, it would be burdensome for the neural network to process all $\hat{M}$ frames, and the contents in adjacent frames could be similar and duplicate each other. Therefore, we compress the voxels by merging nearby frames. For moving scenes, adjacent frames are not completely aligned. Hence methods such as \cite{est,cannici2020matrix} that conduct pixel-wise compression suffer from blurs, lacking a mechanism for the events to communicate in spatial. 

% Justification: Conv to map events with a spatial offset, potentially align
To tackle the misalignment problem during compression, we propose using convolutional layers to aggregate voxels within local spatial-temporal regions. The trainable convolutional layers can project the voxel at $(t, x+\Delta x, y+\Delta y)$ to $(t', x, y)$ on the target frame, with a spatial offset. Hence, the voxels corresponding to the same real-world point can be potentially aligned to the same pixel on the anchor frame.

% Description: Group, Concat on channel, Compress channel
As shown in the right of Fig.~\ref{fig:AET_flow}, for each step, the frames are grouped with a group size $G$, and concatenated on channel, resulting in $M^*=\hat{M}/G$ groups (or new frames) with $G\times C_{in}$ channels. Then they pass through a shared Conv2d which compresses the channels to $1 \times C_{out}$, meaning that within each group, the original $G$ frames with $C_{in}$ channels each are now compressed into $1$ frame with $C_{out}$ channels. This Conv2d operation is equivalent to a strided Conv3d to merge frames, but is better accelerated by PyTorch~\cite{paszke2019pytorch}.

% Justification of AET, from visualization and Ablation study
In Fig.~\ref{fig:AET_vis}, we show the comparison between our AET and EST~\cite{est}. Since this work focuses on end-to-end training, we do not explicitly impose extra loss function to train for sharper images. However, we observed that AET generates sharper outputs with clearer textures. We hypothesis the spatial-temporal compression in AET leaves the room for the model to auto-align the events, and hence learn a sharper representation for higher classification accuracy. In Sec.~\ref{sec:aet_ablation}, the ablation studies also verify the effectiveness of AET and its sub-modules respectively.

\subsection{Event Frame Net}
The Event Frame Net (EFN) processes the AET encoded event data for classification tasks in three steps, CNN encoding, Frame- and Video-branch classification, and Result synthesis. The EFN is designed to be versatile under both static and dynamic scenes. In our ablation study, EFN is verified to be able to boost accuracy effectively.

\begin{figure*}
    \begin{center}
        \includegraphics[scale=0.7]{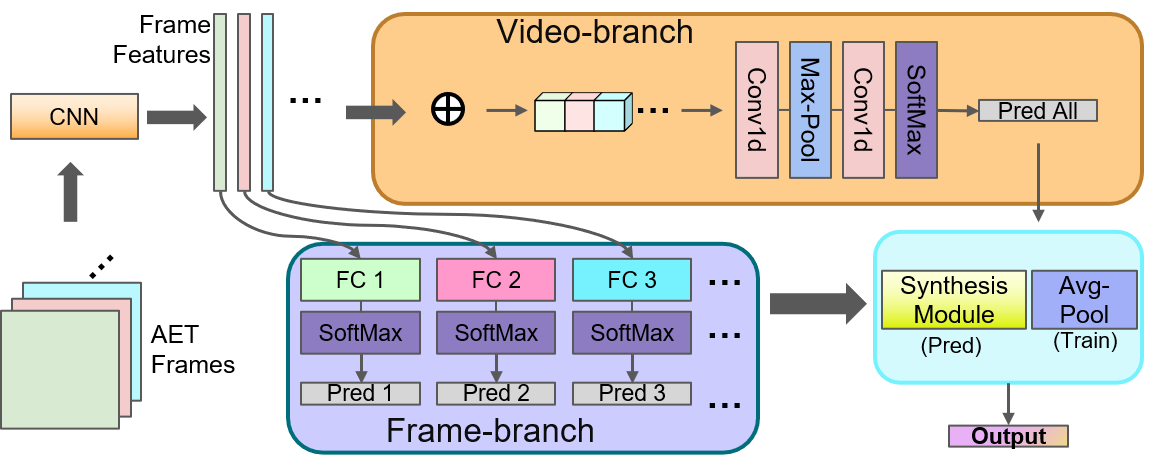}
        \caption{Event Frame Net workflow. The AET frames are encoded into vector representations by the backbone CNN, and then processed in two branches. Frame-branch makes a prediction for each frame with independent classifiers. Video-branch treats the vectors as a sequence and process it with 1D convolutional layers. During training, the predictions are averaged as the final output. For inference, a synthesis module described in Fig.~\ref{fig:synthese_pred} and Sec.~\ref{sec:synthesis_module} is used to merge the results.}
        \vspace{-20pt}
        \label{fig:EFN_workflow}    
    \end{center}
\end{figure*}

\subsubsection{CNN Encoder}
AET encodes event data into tensors of shape $C\times M^* \times H \times W$, denoting $M^*$ frames with $C$ channels each. For each individual frame, we use a shared CNN backbone model to extract geometrical information into a vector. Following \cite{est} and \cite{cannici2020matrix}, we use the ImageNet-pretrained ResNet~\cite{he2016deep} with the last decision-making layer removed.

\subsubsection{Frame branch}
As shown at the bottom of Fig.~\ref{fig:EFN_workflow}, the frame branch generates per-frame predictions with an independent fully-connected layer for each individual frame. In practice, we use a grouped $1\times1$ convolutional layer with $g=M^*$ groups to facilitate parallel computation. Also, a sliding window is usually used for real-time event-based vision~\cite{yang2019gumbel}, so that the number of frames, $M^*$, can be fixed and made compatible with our design under different scenarios. In fact, one may also use a shared classifier for smaller model size, but a 0.2\% accuracy drop was observed.

For static scenes, the frames are similar and contain almost all the information needed for decision making. Additionally, for dynamic tasks such as human action recognition, one may also distinguish different actions by the body figure from just one frame. Therefore, the frame-based classifier can perform well under static and even some dynamic scenes.

\subsubsection{Video branch}
For certain tasks such as moving direction recognition, it is critical to make decision based on a series of frames, rather than individuals. Therefore, we propose the video-branch of EFN, illustrated in the upper part of Fig.~\ref{fig:EFN_workflow}, where the frame representations are concatenated into a sequence and then Conv1d modules generate the video-level prediction.

Techniques such as LSTM~\cite{lstm} and Transformer~\cite{attention_all_you_need} are usually considered more powerful for sequential data processing. However, from experiment results, we found that Conv1d can also achieve similar accuracy with much fewer parameters. We conjecture the sequence length $M^*$ is short enough so that Conv1d can also handle well.

\begin{figure}
    \centering
    \includegraphics[scale=0.42]{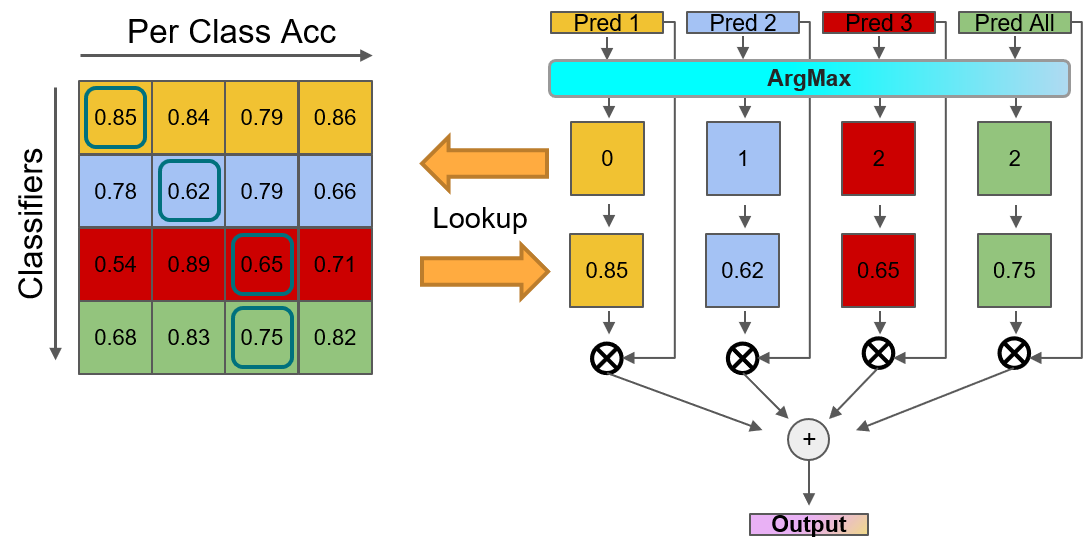}
    \caption{Synthesis Module. The results are merged by a weighted sum, where weights are derived from each classifier's performance on the validation set.}
    \vspace{-15pt}
    \label{fig:synthese_pred}
\end{figure}

\subsubsection{Result Synthesis}
\label{sec:synthesis_module}
To merge the multiple predictions from video-branch and frame-branch, a separate Synthesis module is used during inference of the EFN. The module records per-class accuracy in the validation set, and then merges the predictions with a weighted sum, as shown in Fig.~\ref{fig:synthese_pred}.

Intuitively, since EFN is proposed to handle tasks under both static and dynamic scenes, the reliability of classifiers in the frame-branch and video-branch may also fluctuate correspondingly. Subsequently, we employ a statistical method to weight these predictions. We evaluate the trained model on the validation set while recording the per-class accuracy $Acc(p, q)$, which is the accuracy when the classifier $p$ makes a prediction of class $q$. The per-class accuracy is stored as a matrix shown on the left of Fig.~\ref{fig:synthese_pred}.

The final output of the synthesis module is derived by a weighted sum of the prediction logits. Firstly, each classifier generates a prediction vector, indicating a class via ArgMax. Then the per-class accuracy is queried, with keys being the classifier and its prediction. The per-class accuracies are used as the weights for weighted-summing the classifiers' predictions, which can be formulated as:
\begin{equation}
\textrm{out} = \sum_{p}{Acc(p, \textrm{argmax}(l_p)) \times l_p}
\end{equation}
where $l_p$ is the prediction logits from classifier $p$. 

\section{Experiments}
\begin{table*}[ht]
    \centering
    \begin{tabular}{|l|l|ccccc|}
    \hline
    Dataset      & Type    & \#classes & Total samples & Avg. events (kEv) & Avg. length (s) & Avg kEv/s \\ \hline
    \hline
    N-Caltech101 & Static  & 101         & 8,709         & 112.76            & 0.30            & 375.24          \\
    N-Cars       & Static  & 2           & 24,029        & 3.09              & 0.10            & 30.93           \\
    N-Caltech256 & Static  & 257         & 30,607        & 104.96            & 1.00            & 105.65          \\
    DVS128       & Dynamic & 11          & 1,464         & 367.98            & 6.52            & 55.62           \\
    UCF50        & Dynamic & 50          & 6,681         & 1030.83           & 6.80            & 160.25         \\ \hline
    \end{tabular}
    \vspace{-5pt}
    \caption{Statistics of the datasets used in the experiments.}
    \vspace{-10pt}
    \label{table:datasets}
\end{table*}

Detailed experiments are conducted on our approach to evaluate the overall performance, each module’s contribution and the efficiency. Our method consistently surpasses all baselines under both static and dynamic scenes.

\subsection{Datasets and Preprocessing}
We evaluate our approach on datasets of static and dynamic scenes, whose statistics are shown in Table~\ref{table:datasets}. The N-Caltech101~\cite{ncal101} and N-Cars~\cite{sironi2018hats} are benchmark datasets used in many previous studies, such as~\cite{est, cannici2020matrix, sironi2018hats}. They contain static daily objects and cars. Besides, recently Ref.~\cite{dvsDatasets} proposes a new static dataset N-Caltech256 with many more samples and classes, and therefore more challenging. Hence, we believe consistent improvements on this dataset would be solid and convincing.

For dynamic scenes, the DVS128 dataset~\cite{dvs128} records human gestures under different illumination environments, which has been adopted in \cite{yang2019gumbel}. A new dynamic and challenging dataset, the UCF-50 in event vision, is also newly proposed by \cite{dvsDatasets}, which contains sports actions.

The datasets are split into training, validation, and test sets following official practices. The DVS128 dataset is preprocessed by cutting the event flow with a window size of 750ms and 100ms steps according to~\cite{yang2019gumbel}. Among them, the N-Caltech101, N-Cars, and DVS128 are recorded in the real-world, while the other datasets are converted from conventional RGB datasets by simulators.

\subsection{Experimental Settings}
We train our network on the five datasets with the Adam optimizer at a learning rate of 1e-4. The learning rate is adjusted step by step with a cosine scheduler, with a 10\% linear warm-up. The model is trained for 200 epochs for N-Caltech256 and 120 epochs for other datasets.

For AET, we quantize the timestamps with $\hat{M}=100$ and use 2-step Conv2d for the compression. Kernel size is 5, $G=10$ (2, 5 respectively) and LeakyReLU is used for activation. $C_{in}=1, C_{out}=3$, and intermediate channel = 4. For the EFN video-branch, 2 convolutional layers with kernel size $k_1=5, k_2=3$ are used. We also use a max-pooling layer of size 2, resulting in 1 element in the end. By default, Res18 is used for the backbone.

In the training process, predictions from different classifiers are averaged and CrossEntropyLoss is used. The best model with the highest average accuracy on the validation set is selected for testing. During inference, the Synthesis module is used to combine the predictions.

\subsection{Results on Static Datasets}
\begin{table}[]
\centering
\resizebox{\linewidth}{!}{
\begin{tabular}{|l|l|c|c|c|}
\hline
Model                                             & Backbone & N-Cars         & N-Cal101       & N-Cal256       \\ \hline
HOTS~\cite{lagorce2016hots}                        & -    & 62.4           & 21.0           & -              \\
HATS~\cite{sironi2018hats}                         & Res34    & 90.9           & 69.1           & -              \\
Two-Channel~\cite{twoChannel}                      & Res34    & 86.1           & 71.3           & -              \\
Voxel Grid~\cite{voxelGrid}                        & Res34    & 86.5           & 78.5           & -              \\
EST~\cite{est}                                    & Res34    & 92.5           & 83.7           & 63.11          \\
Matrix-LSTM~\cite{cannici2020matrix}               & Res18    & \textit{94.37}          & \textit{84.31}          & -              \\ 
\hline
MLSTM+E2VID\cite{cannici2020matrix}               & Res34    & (95.65)          & (85.72)          & -              \\
MLSTM+E2VID\cite{cannici2020matrix}               & Res18    & (95.80)          & (84.12)          & -              \\
\hline
{AET-EFN (Ours)} & Res18    & \textbf{95.91} & \textbf{89.25} & \textbf{67.86} \\
{AET-EFN (Ours)} & Res34    & \textbf{95.28} & \textbf{89.95} & \textbf{68.51} \\
\hline
\end{tabular}}
\vspace{-5pt}
\caption{Accuracy (\%) on static datasets. Second best in italic. Our method performs strongly even with the lighter Res18. MatrixLSTM+E2VID~\cite{cannici2020matrix} is reconstruction-based hence not in the scope of this research, which focuses on end-to-end training, but we still surpass their results.}
\label{table:static_results}
\vspace{-10pt}
\end{table}

As shown in Table~\ref{table:static_results}, our proposed method surpasses current methods by a large margin, including HOTS~\cite{lagorce2016hots}, HATS~\cite{sironi2018hats}, Two-Channel Image~\cite{twoChannel}, Voxel Grid~\cite{voxelGrid}, Matrix-LSTM~\cite{cannici2020matrix} and EST~\cite{est}. Our approach achieves the SOTA performance in both static N-Caltech101 and N-Cars. Compared with the second-best solution, AET-EFN increases the prediction accuracy by 1.54\% and 5.64\%, respectively. 

The N-Caltech256 is a new dataset that has not been used in previous works. We re-implemented EST~\cite{est} on N-Caltech256 as the baseline, because EST is among the best methods on the previous two datasets and is open-sourced. With the new dataset, AET-EFN can still improve the accuracy by 5.4\% over EST. Considering the baselines can benefit from the large amount of data, the improvement in N-Caltech256 strongly proves the superiority of our method.

Moreover, as in Table~\ref{table:CNN_backbone} and Sec~\ref{sec:cnn_ablation}, we found that more powerful backbones can further boost the accuracy, even with fewer parameters. Specifically, AET-EFN can achieve an impressive 91.95\% accuracy on the N-Caltech101 dataset using an EfficientNet backbone~\cite{tan2019efficientnet}.

\subsection{Results on Dynamic Datasets}
\begin{table}[t]
    \centering
    \begin{tabular}{|l|c|cc|}
    \hline
    Model         & Type   & DVS128 & UCF50 \\ \hline
    \hline
    Amir \etal \cite{dvs128}& PointCloud & 94.4   & -     \\
    PointNet~\cite{qi2017pointnet}      & PointCloud & 88.8   & -     \\
    PointNet++~\cite{qi2017pointnet++}    & PointCloud & 95.2   & -     \\
    PAT~\cite{yang2019gumbel}           & PointCloud & 96.0   & -     \\
    EST~\cite{est} (Res18)         & Voxel     & 96.91  & 67.99 \\ \hline
    AET-EFN (Res18)   & Voxel     & \textbf{97.38}  & \textbf{84.03} \\
    \hline
    \end{tabular}
    \vspace{-5pt}
    \caption{Accuracy (\%) on dynamic datasets.The methods listed in the table are either point-cloud-based or voxel-based. Res18 backbone is used for both EST and AET-EFN.}
    \vspace{-10pt}
    \label{table:dynamic_results}
\end{table}
Table~\ref{table:dynamic_results} shows the results on dynamic datasets, wherein the proposed method outperforms all baselines and improves the accuracy to 97.38\% for DVS128 and 84.03\% for UCF-50. It surpasses point-cloud-based methods, such as Amir \etal~\cite{dvs128}, PAT~\cite{yang2019gumbel}, PointNet~\cite{qi2017pointnet} and PointNet++~\cite{qi2017pointnet++}. While the methods in Table~\ref{table:static_results} only consider static scenes and have not been applied to dynamic datasets, we re-implemented the EST method~\cite{est} as the representative of the voxel-based class. Surprisingly, the EST method achieves an accuracy of 96.91\%. We hypothesize since DVS128 contains obviously different gesture classes with little noise, it could be an easy task. Moreover, the recordings have been clipped into short windows~\cite{yang2019gumbel}, hence the motion blur problem is also mitigated. Nonetheless, our approach still achieves a new SOTA accuracy on DVS128.

The UCF-50 is also a new dataset on which no existing work has been applied. Different from DVS128, the dataset contains much longer recordings and complex actions, hence less reasonable for clipping. The point-cloud-based methods are not applicable due to the cloud size (1030k events per sample on average) and noise. Therefore, we re-implemented the EST, which is also the best baseline for DVS128, on UCF-50. For this challenging dataset, EST drops to 67.99\%, whereas our method achieves 84.03\%, viz. a substantial improvement of 16.04\%.

\subsection{Ablation Study}
We conduct ablation studies to analyze the contributions of different modules in the following.

\begin{table}[t]
\centering
\begin{tabular}{|l|c|c|}
\hline
Backbone CNN & \#params (Mb) & NCal101 Acc (\%)                       \\ \hline
\hline
Res18~\cite{he2016deep}        & 11.18        & 89.25                           \\
Res34~\cite{he2016deep}        & 21.28        & 89.95                           \\
Dense121~\cite{densenet}     & 7.48         & 90.12                           \\
ENet\_lite0~\cite{tan2019efficientnet}  & 4.03         & 90.75                           \\
ENetB2~\cite{tan2019efficientnet}       & 8.42         & \textbf{91.95}                  \\ \hline
\end{tabular}
\vspace{-5pt}
\caption{Backbone ablation study on N-Cal101.}
\vspace{-15pt}
\label{table:CNN_backbone}
\end{table}

\subsubsection{Backbone CNN}
\label{sec:cnn_ablation}
The EFN adopts a CNN backbone as the encoder to learn a vector representation for each frame of AET. Previous tensor-based methods, such as~\cite{est,cannici2020matrix,sironi2018hats}, adopt the popular ResNet~\cite{he2016deep} as their backbone. To study the effect of different CNN models on the accuracy of EFN, we run ablation studies on the N-Caltech101 dataset. As shown in Table~\ref{table:CNN_backbone}, EFN benefits from more powerful CNN backbones consistently. Compared with EFN + Res18 configuration, the adoption of the EfficientNet\_B2~\cite{tan2019efficientnet} increases the accuracy by 2.7\% to 91.95\%. With similar or smaller model size. Other choices, such as EfficientNet\_lite and DenseNet~\cite{densenet} can also improve the accuracy. These results confirm that the improvement brought by AET-EFN is universal. Even better, the EFN framework is orthogonal to its backbone and can harvest from more advanced CNN models.

\begin{table}[t]
\begin{center}
\begin{tabular}{|l|c|c|}
\hline
Method      & Variant       & NCal101 Acc (\%)       \\ \hline
\hline
AET + EFN   & -               & \textbf{89.25} \\ 
AET + Res18 & -               & 86.90          \\
EST\cite{est} + EFN & -       & 86.50          \\ 
\hline
            & Spike           & 88.57          \\
AET variants   & Spike+Accum     & 88.22          \\
 + EFN            & Avg compress    & 87.19          \\
            & Quantize only & 86.61          \\ \hline
AET & no video-branch & 88.45         \\
 +EFN variants           &  no frame-branch & 87.24         \\ \hline

\end{tabular}
\end{center}
\vspace{-10pt}
\caption{Ablation study on AET and EFN, the results prove that both AET and EFN can bring significant improvements independently. For AET variations, `Spike' means directly voxelizing the quantized events without Accumulative Voxelization. `Avg compress' averages adjacent frames for compression instead of the original Aligned Compression. `Quantize only' quantizes the events into 10 frames directly without the later procedures. Discussion in \ref{sec:aet_ablation}.}
\vspace{-5pt}
\label{table:AET_EFN_ablation}
\end{table}

\subsubsection{Ablation on AET}
\label{sec:aet_ablation}
To quantify the contributions of the AET module and its three sub-modules, we conduct ablation studies in Table~\ref{table:AET_EFN_ablation}. Compared with the original AET+EFN design (89.25\%), replacing the AET module with EST~\cite{est} degrades the accuracy by 2.75\% to 86.50\%, showing the significance of the AET module.

Moreover, in the `AET variants' part, we show the contribution of the AET sub-modules. In this part, except the specific changes, other settings are the same as AET. The last row `Quantize only' directly quantizes events into 10 bins without the two sub-modules, Accumulative Voxelization and Aligned Compression, resulting in a 2.64\% drop. The first row `Spike' omits the Accumulative Voxelization step. Hence each voxel represents spikes of the events rather than accumulative intensity changes. The `Spike+Accum' denotes a hybrid of the two voxelization methods. However, both of them decrease the accuracy by 0.68\% and 1.03\%, respectively, showing the contribution of the Accumulative Voxelization. Finally, the `Avg compress' in the third row replaces the Aligned Compression by averaging the frames in groups, where the accuracy also decreases by 2.06\%.

\subsubsection{Ablation on EFN}
Next, we also study the contribution of the EFN framework. As listed in Table~\ref{table:AET_EFN_ablation}, the accuracy drops by 2.35\% to 86.90\% without the EFN. As for the two branches in EFN, the accuracies are 0.8\% and 2.01\% lower if the video-branch and the frame-branch are removed, respectively. These results indicate that EFN can effectively improve the accuracy and both branches contribute to the improvement.

\subsection{Influence of Hyper-Parameters}
\label{sec:hyper_param}
To study the impact of the hyper-parameters $\hat{M}$ and $G$, we conduct experiments on N-Caltech101 as in Table~\ref{tab:hyperparam}. According to the results, the choice of $\hat{M}$ for Timestamp Quantization and $G$ for Auto-aligned Compression has only little impact on the final accuracy, within a proper range. Except when the group size equals 50, the accuracy drops by 1.2\%, which is probably due to the information loss caused by the high compression rate. We also test different kernel sizes (K) for the Auto-Aligned Compression in AET. A larger kernel appears to benefit the prediction, which is reasonable since it implies a larger receptive field for alignment.

\begin{table}[t]
    \centering
\begin{tabular}{|c|c|c|c|c|c|c|c|}
\hline
$\hat{M}$ & Acc(\%) &  & G & Acc(\%) &  & K           & Acc(\%) \\ \hline
50                 & 88.86   &  & 5      & 89.25   &  & 3 & 88.23   \\
100                & 89.25   &  & 10     & 89.25   &  & 5 & 89.25   \\
150                & 89.26   &  & 20     & 88.91   &  & 9 & 89.43   \\
200                & 88.91   &  & 50     & 88.05   &  &   &    \\ \hline
\end{tabular}
    \vspace{-5pt}
    \caption{Impact of hyper-parameters on N-Caltech101, detailed discussion in Sec.~\ref{sec:hyper_param}}
    \vspace{-5pt}
    \label{tab:hyperparam}
\end{table}

\subsection{Efficiency of our approach}
\begin{table}[t]
\centering
\resizebox{\linewidth}{!}{
\begin{tabular}{|l|c|cc|}
\hline
Method      & Asynch. & Time (ms)     & Speed (kEv/s)   \\ \hline
\hline
Gabor-SNN~\cite{sironi2018hats}   & Yes     & 285.95        & 14.15           \\
HOTS~\cite{lagorce2016hots}      & Yes     & 157.57        & 25.68           \\
HATS~\cite{sironi2018hats}        & Yes     & 7.28          & 555.74          \\ 
Matrix-LSTM~\cite{cannici2020matrix}& No      & 8.25          & 468.36          \\ 
EST~\cite{est}        & No      & \textit{6.26} & \textit{632.9}  \\ \hline
AET-EFN (Res34) & No      & \textbf{4.64} & \textbf{933.51} \\
AET-EFN (Res18) & No      & \textbf{3.18}     & \textbf{1194.20}         \\ \hline

\end{tabular}
}
\vspace{-5pt}
\caption{Inference Speed of AET-EFN. The second column indicates whether the method processes new events asynchronously or synchronously in packages. For fairness, we list the results for both Res18 and Res34. NCars is used following the baseline methods.}
\vspace{-15pt}
\label{table:Speed}
\end{table}

Besides accuracy improvement, the proposed method also outperforms other methods in terms of efficiency. Following~\cite{est, cannici2020matrix}, we use the inference time and the average event processing speed as metrics. We use one GTX 1080Ti GPU and predict one sample at a time, same as in \cite{cannici2020matrix} for fairness. The methods are categorized into asynchronous and synchronous classes. The asynchronous methods, such as Gabor-SNN~\cite{sironi2018hats}, HOTS~\cite{lagorce2016hots} and HATS~\cite{sironi2018hats} process the input events asynchronously upon their arrival. On the other hand, synchronous methods such as Matrix-LSTM~\cite{cannici2020matrix} and EST~\cite{est} process the events in packets.

From Table~\ref{table:Speed}, the efficiency of AET-EFN is the highest among its fellows. Following \cite{est,cannici2020matrix}, we conduct the efficiency test on the N-Cars dataset. To fairly evaluate the speed of EFN, we use Res34 as the CNN backbone, same as other baselines. Benefiting from the efficient AET representation, our approach achieves the lowest inference latency and fastest event processing speed, at 4.64ms and 933.51kEvents/s, respectively. Moreover, with the Res18 backbone, the AET-EFN is even faster at 3.18ms, equivalent to an impressive 314Hz frame rate.

\section{Visualization}
\begin{figure}
    \centering
    \begin{subfigure}[]{0.24\columnwidth}
        \includegraphics[width=\linewidth]{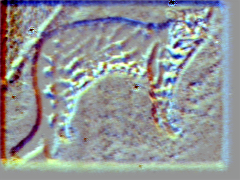}
    \end{subfigure}
    \begin{subfigure}[]{0.24\columnwidth}
        \includegraphics[width=\linewidth]{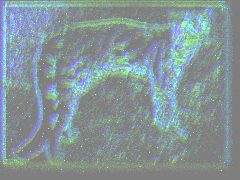}
    \end{subfigure}
    \begin{subfigure}[]{0.24\columnwidth}
        \includegraphics[width=\linewidth]{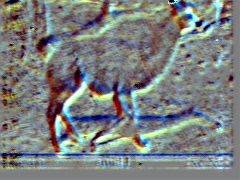}
    \end{subfigure}
    \begin{subfigure}[]{0.24\columnwidth}
        \includegraphics[width=\linewidth]{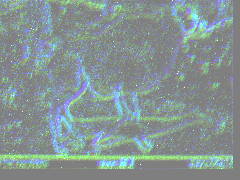}
    \end{subfigure}
    % \begin{subfigure}[]{0.3\columnwidth}
    %     \includegraphics[width=\linewidth]{complex2.png}
    % \end{subfigure}
    % \begin{subfigure}[]{0.3\columnwidth}
    %     \includegraphics[width=\linewidth]{complex2a.png}
    % \end{subfigure}
    
    % \newline
    
    \begin{subfigure}[]{0.24\columnwidth}
        \includegraphics[width=\linewidth]{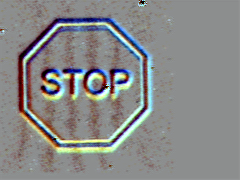}
    \end{subfigure}
    \begin{subfigure}[]{0.24\columnwidth}
        \includegraphics[width=\linewidth]{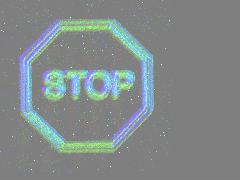}
    \end{subfigure}
    \begin{subfigure}[]{0.24\columnwidth}
        \includegraphics[width=\linewidth]{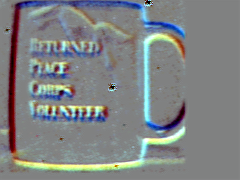}
    \end{subfigure}
    \begin{subfigure}[]{0.24\columnwidth}
        \includegraphics[width=\linewidth]{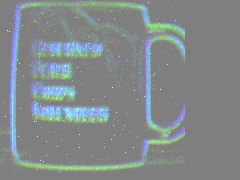}
    \end{subfigure}
    % \begin{subfigure}[]{0.3\columnwidth}
    %     \includegraphics[width=\linewidth]{text2.png}
    % \end{subfigure}
    % \begin{subfigure}[]{0.3\columnwidth}
    %     \includegraphics[width=\linewidth]{text2a.png}
    % \end{subfigure}
    
    % \newline
    
    \begin{subfigure}[]{0.24\columnwidth}
        \includegraphics[width=\linewidth]{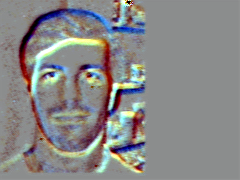}
    \end{subfigure}
    \begin{subfigure}[]{0.24\columnwidth}
        \includegraphics[width=\linewidth]{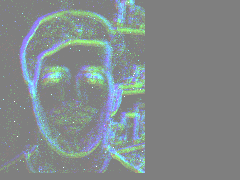}
    \end{subfigure}
    \begin{subfigure}[]{0.24\columnwidth}
        \includegraphics[width=\linewidth]{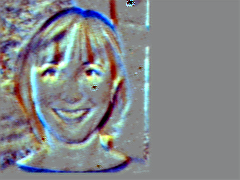}
    \end{subfigure}
    \begin{subfigure}[]{0.24\columnwidth}
        \includegraphics[width=\linewidth]{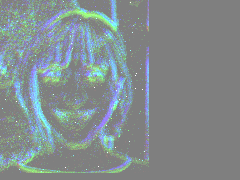}
    \end{subfigure}
    % \begin{subfigure}[]{0.3\columnwidth}
    %     \includegraphics[width=\linewidth]{human2.png}
    % \end{subfigure}
    % \begin{subfigure}[]{0.3\columnwidth}
    %     \includegraphics[width=\linewidth]{human2a.png}
    % \end{subfigure}
    
    % \newline
    \begin{subfigure}[]{0.24\columnwidth}
        \includegraphics[width=\linewidth]{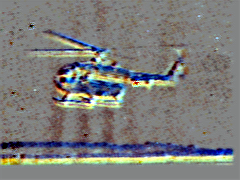}
    \end{subfigure}
    \begin{subfigure}[]{0.24\columnwidth}
        \includegraphics[width=\linewidth]{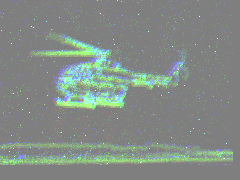}
    \end{subfigure}
    \begin{subfigure}[]{0.24\columnwidth}
        \includegraphics[width=\linewidth]{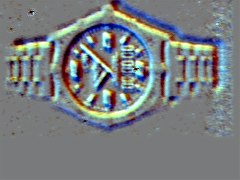}
    \end{subfigure}
    \begin{subfigure}[]{0.24\columnwidth}
        \includegraphics[width=\linewidth]{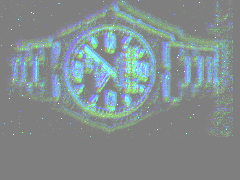}
    \end{subfigure}
    % \begin{subfigure}[]{0.3\columnwidth}
    %     \includegraphics[width=\linewidth]{obj2.png}
    % \end{subfigure}
    % \begin{subfigure}[]{0.3\columnwidth}
    %     \includegraphics[width=\linewidth]{obj2a.png}
    % \end{subfigure}
    
    \vspace{-5pt}
    \caption{Visualizations of AET (left) and EST~\cite{est} (right) in pairs. Our method enjoys sharper edges, clearer texture and finer details.}
    \vspace{-10pt}
    \label{fig:AET_vis}
\end{figure}
In Fig.~\ref{fig:AET_vis} we visualize our AET in comparison with the EST representations. Following~\cite{est}, we take the average of the channels to reduce them to 3 for RGB display. It is commonly observed that AET enjoys perceptibly better quality images, which provides higher interpretability to our method (e.g., under noisy background in the first row and clearer human faces in the third row). Moreover, differences are even more obvious when there are texts, as shown in the second row. Some interesting patterns, such as affluent contrast and sharp edges have been observed, which further confirms the effectiveness of the proposed AET. Together with the results in Sec.~\ref{sec:aet_ablation}, we believe this visualization can justify our AET as a better representation. 

\section{Conclusion}
This work has proposed the Aligned Event Tensor (AET) and Event Frame Net (EFN) for event-based vision under both static and dynamic scenes. Extensive experiments have confirmed the integration of AET and EFN consistently outperforms all other SOTA methods by large margins. The contributions of AET and EFN have been extensively quantified through ablation studies. The AET appears to be a generically better voxel-based representation for event data under different scenes, and may be adopted for various event-vision tasks. The versatile EFN also improves the prediction accuracy further. Compared with existing approaches, AET-EFN is provably superior in inference time and event processing speed, rendering it promising for real-time online tasks.

{\small
\bibliographystyle{ieee_fullname}
\bibliography{references}
}

\end{document}